\documentclass[conference]{IEEEtran}
\usepackage{doi}

\usepackage{cite}
\usepackage{amsmath,amssymb,amsfonts}
\usepackage{algorithmic}
\usepackage{graphicx}
\usepackage{textcomp}
\usepackage{xcolor}
\usepackage{booktabs}
\usepackage{hyperref}
\def\BibTeX{{\rm B\kern-.05em{\sc i\kern-.025em b}\kern-.08em
    T\kern-.1667em\lower.7ex\hbox{E}\kern-.125emX}}
\begin{document}

\title{Localized Uncertainty Quantification in Random Forests via Proximities}


\author{\IEEEauthorblockN{1\textsuperscript{st} Jake S. Rhodes}
\IEEEauthorblockA{\textit{Department of Statistics} \\
\textit{Brigham Young University}\\
Provo, Utah, USA \\
rhodes@stat.byu.edu}
\and
\IEEEauthorblockN{2\textsuperscript{nd} Scott D. Brown}
\IEEEauthorblockA{\textit{Department of Statistics} \\
\textit{Brigham Young University}\\
Provo, Utah, USA \\
scottdbrown7@gmail.com}
\and
\IEEEauthorblockN{3\textsuperscript{rd} J. Riley Wilkinson}
\IEEEauthorblockA{\textit{Department of Statistics} \\
\textit{Texas A\&M University}\\
College Station, Texas, USA \\
j.riley.wilkinson@gmail.com}
}

\maketitle

\begin{abstract}
    In machine learning, uncertainty quantification helps assess the reliability of model predictions, which is important in high-stakes scenarios. Traditional approaches often emphasize predictive accuracy, but there is a growing focus on incorporating uncertainty measures. This paper addresses localized uncertainty quantification in random forests. While current methods often rely on quantile regression or Monte Carlo techniques, we propose a new approach using naturally occurring test sets and similarity measures (proximities) typically viewed as byproducts of random forests. Specifically, we form localized distributions of OOB errors around nearby points, defined using the proximities, to create prediction intervals for regression and trust scores for classification. By varying the number of nearby points, our intervals can be adjusted to achieve the desired coverage while retaining the flexibility that reflects the certainty of individual predictions. For classification, excluding points identified as unclassifiable by our method generally enhances the accuracy of the model and provides higher accuracy-rejection AUC scores than competing methods.
\end{abstract}


\section{Introduction}

Although traditional machine learning models usually provide point estimates, there is growing recognition of the need to incorporate uncertainty to support more informed decisions~\cite{abdar2021review}. By quantifying uncertainty, users can assess the reliability of model outputs and better interpret results, especially for out-of-distribution samples through calibrated confidence estimates.

Global measures, such as mean squared error (MSE), overall accuracy, precision, or recall, are usually used to quantify the quality of machine learning models. However, additional insights into model performance can be provided by assessing local or instance-based contributions~\cite{rosaler2023enhanced}. This involves considering the likelihood of prediction accuracy based on similarities with other observations, thus providing a more nuanced understanding of model reliability. In doing so, users can identify regions of the data in which high or low levels of confidence in the predicted values should be placed.

In this work, we focus on the uncertainty quantification of random forests. We use a naturally occurring test set that is often viewed as a byproduct of random forests to determine an error distribution that we can extend to new points. Localized error distributions are used to generate prediction intervals (for regression problems) or trust scores (for classification problems) that provide additional information on the decision-making process.

\subsection{Random Forests}

Random forests were introduced in 2001 by Leo Breiman~\cite{Breiman2001randomforests} and continue to serve as a reliable model in scientific fields. However, like many other machine learning models, random forests are still widely considered black-box models. That is, the path from the data to the prediction is not conveyed in a human-interpretable way. As a partial resolution, the random forest predictions can be described in terms of a weighted sum of responses, as first suggested in~\cite{lin2006adaptiveNN}, and later extended to classification problems in~\cite{rhodes2023rfgap}. From this point of view, the random forest predictions can be broken down into instance-based contributions, where the weights describe how specific training instances influence the prediction.

In~\cite{rhodes2023rfgap}, the authors described these random forest weights as proximities or similarity measures. They demonstrated their superiority over other random forest proximities in several tasks, including data imputation, outlier detection, and visualization~\cite{rhodes2021rfphate}. It was argued that because these proximities could serve as weights capable of reconstructing the random forest predictions, they preserved the data geometry learned by the random forest and are thus called random forest geometry and accuracy-preserving proximities, or RF-GAP. We use the RF-GAP proximities to provide instance-based (localized) measures of uncertainty.


\section{Related Works}

Measures of uncertainty in random forests span several approaches, including variance- or density-based measures at the leaf or tree level, bootstrapping or Monte Carlo resampling, and methods leveraging OOB (OOB) errors or proximities. Most works focus on regression, though classification uncertainty has also been addressed. We summarize these works below.

\subsection{Regression Methods}

Early approaches assessed uncertainty using structural properties of the trees. Dutschmann and Baumann~\cite{dutschmann2021molecules} introduced high-variance leaf nodes, deeming predictions less certain when many such nodes were involved, though this rarely improved over standard deviations of the ensemble. Meinshausen~\cite{meinshausen2006quantileforests} proposed quantile regression forests (QRFs), which extend random forests by estimating conditional distributions rather than conditional means. Instead of storing only node averages, QRFs retain all response values within each terminal node, from which a weighted empirical distribution is constructed. This enables direct estimation of quantiles, allowing for prediction intervals and outlier detection. While QRFs provide flexible, distributional predictions, they do not explicitly account for random forest error distributions.

Bootstrap-based strategies estimate variance through resampling. Wager et al.~\cite{wager2014rfjackknife} developed jackknife and infinitesimal jackknife variance estimates, attributing uncertainty to Monte Carlo and sampling noise, though requiring impractically large resamples. Coulston et al.~\cite{coulston2016approxuncertainty} trained thousands of forests on bootstrap samples, producing well-calibrated intervals but at high computational cost. Hooker and Mentch~\cite{hooker2016rfquantuncertain} applied PAC theory and U-statistics for prediction intervals with guaranteed coverage, though exhaustive computation is infeasible, so incomplete subsampling is required.  

OOB error distributions provide a natural test set for interval construction. Zhang et al.~\cite{zhang2020RFPredInterval} used OOB residuals with conformal analysis to build prediction intervals, which achieve valid asymptotic coverage and are narrower than QRFs, but remain globally uniform and thus less informative at the instance level. Recent work also highlights the role of proximities for local explanation and error attribution, showing that only a small fraction of nearest neighbors contribute meaningfully to predictions~\cite{rosaler2023enhanced}.  

\subsection{Classification Methods}

For classification, uncertainty measures often rely on density or proximity information. Fornaser et al.~\cite{fornaser2018sigma} proposed the sigma-z random forest, estimating uncertainty from split-based probability densities. The method assigns lower confidence to misclassified points but is slow to train and lacks comprehensive evaluation. Proximity-based methods instead exploit neighborhood structure. Devetyarov and Nouretdinov~\cite{devetyrov2010predswconfidence} measured uncertainty by comparing average proximities of same-class versus other-class neighbors, producing effective conformity scores for OOB samples, though extension to unseen points remains unclear. More generally, proximity information can be leveraged to identify heterogeneous decision regions where predictions are less trustworthy.  

Random forests provide information about local relationships between points encoded as weights or proximities. Information from nearby points can be used to supplement the model's prediction. In~\cite{rosaler2023enhanced}, random forest weights are used to provide instance-level model explainability. They show that only a small fraction (15\% in their case) of the nearest neighbors contributed to the actual prediction of their regression problem. Nearest-neighbor errors were attributed to errors in the nearby training points.

Similarly to this approach, we use the distribution of out-of-bag (OOB) errors to approximate test error. Unlike global uncertainty estimates, however, our method builds localized error distributions by incorporating proximity between training and test points. This accounts for varying noise levels across the feature space: a test error estimate is weighted by the closeness of training points. Test points with predictions far from the truth thus receive wider intervals, reflecting nearby OOB errors. 

\section{Random Forest Proximities}

A random forest is a supervised learning model that predicts labels using an ensemble of randomized CART~\cite{breiman1984cart} decision trees. Each tree is trained on a bootstrap sample of the data, with in-bag points used for training and OOB points reserved for internal validation. Trees grow by recursively splitting the data until a stopping criterion is met, resulting in terminal nodes that define the decision space.

In~\cite{lin2006adaptiveNN}, the authors demonstrated that random forest regression models are equivalent to a nearest neighbor problem with an adaptive local metric. The RF-GAP~\cite{rhodes2023rfgap} proximities defined this measure in a way that extended its utility to OOB predictions in both regression and classification settings. For a training pair $(x_i, x_j)$, the RF-GAP proximity is defined as:
\[
w_{ij} = p(x_i, x_j) = \frac{1}{|S_i|} \sum_{t \in S_i} \frac{I(x_j \in J_i(t)) \cdot c_j(t)}{|M_i(t)|}
\]
where $S_i$ is the set of trees where $x_i$ is OOB; $J_i(t)$ is the multiset of in-bag points sharing a terminal node with $x_i$ in tree $t$; $c_j(t)$ is the count of $x_j$ in the bootstrap sample for $t$; and $|M_i(t)|$ is the size of $J_i(t)$. RF-GAP measures the co-occurrence of $x_i$ and $x_j$, weighted by the frequency of $x_j$ in the bootstrap and normalized by the node size. Importantly, we note that ($w_{ii} = 0$, for all indices $i$), which allows the proximities to be useful for applications on OOB training points. 

OOB and test predictions—both in regression and classification—can be computed using RF-GAP proximities as weights in a weighted sum or vote, attributing predictions to nearby training instances. The random forest OOB or test prediction for $x_i$, is defined to be

\[\hat{y}_i=\sum_{j=1}^n w_{ij} y_j\]

This aligns the model output with the influence of neighboring points. RF-GAP has outperformed traditional proximities in various applications~\cite{rhodes2023rfgap}, motivating its use in our uncertainty applications.

\section{Methods}

Instance-level attribution, as defined by RF-GAP proximities,  provides a way to understand how nearby points influence not only a prediction but also the confidence in that prediction. The relationships captured by random forest proximities link test errors to OOB errors, offering insight into where test points fall within the broader distribution of model errors. By leveraging this relational structure, we can derive meaningful measures of uncertainty or confidence in the model’s outputs.

Since OOB observations are not used to train a given decision tree, they serve as a proxy for unseen data, simulating what the model may encounter when deployed. The model's performance on OOB points provides insights into the model's handling of new data. For example, the distribution of OOB errors can reasonably model expected test error rates, assuming that training and test data arise from the same generative model. This assumption was applied to prediction intervals in~\cite{zhang2020RFPredInterval}. Test points close in proximity to OOB points will likely exhibit similar behavior; it is reasonable to assume that the test errors will resemble a local subset of the OOB error distribution. Thus, we can estimate the model's performance on these test points by leveraging test-point proximities to OOB points. 

For simplicity in describing our methods, we denote the RF-GAP weight matrix as $\mathbf{W}$, which can be formed between training points or between training and test points, with $\mathbf{W} \in \mathbb{R}^{n \times n}$ when working with OOB predictions, or $\mathbf{W} \in \mathbb{R}^{n \times n_{\text{test}}}$ for test prediction uncertainty measures.

We present locally defined uncertainty measures based on RF-GAP proximities and the OOB error distribution for regression and classification tasks. The first method (Section~\ref{sec:regression}) targets regression forests and constructs flexible prediction intervals, comparable to quantile regression forests and OOB intervals~\cite{zhang2020RFPredInterval}. The second method (Section~\ref{sec:classification}) is for classification forests and includes two variations to identify trustworthy classifications based on a user-defined threshold.

\subsection{Regression - Prediction Intervals}\label{sec:regression}


When predicted values deviate from true responses, we expect the errors for similar points to have similar direction and magnitude. Using RF-GAP proximities, predictions can be broken down into weighted observational contributions. Thus, we attribute expected error rates at the point level by leveraging the random forest structure to quantify the similarity between the test and training points, resulting in more accurate point-wise error estimates. If the OOB predictions for training points are close to those for a new test point $x_0$, then we expect, on average, a similar level of accuracy for $x_0$.

One approach to estimating local expectation is to set up a proximity-weighted sum of OOB errors of nearby points, as given by

\[
\hat{y}_0 \pm \sum_{i = 1}^{n} {w_{0i} L(y_i, \hat{y}_i)},
\]

where $w_{0i}$ is the RF-GAP proximity between $x_0$ and $x_i$, and $L(., .)$ denotes the loss function used to generate the decision trees, such as MSE. Under this approach, test points with larger errors will, on average, have larger interval widths as determined by the locally weighted expected OOB errors of nearby training points. These proximity-weighted errors provide a meaningful context for uncertainty; however, the method does not convey any notion of expected coverage. Therefore, to create prediction intervals, we gauge the expected error of $x_0$ by estimating the quantiles of the OOB error distribution in a local neighborhood defined by the RF-GAP proximities.

Our proposed method of constructing prediction intervals starts by defining localized distributions of the OOB training errors. We call our method Random Forest Interval Residual Estimation, or RF-FIRE. Similarly to the intervals described in~\cite{zhang2020RFPredInterval}, RF-FIRE is based on OOB errors calculated from the trained random forest. Since these OOB predictions constitute a natural validation set---assuming that any new data follow the same predictor-response relationship as the training data--the functional form of the unknown residuals of the new data should also match, following the principles of conformal analysis~\cite{vovk2005algorithmic}. This idea was exploited in~\cite{zhang2020RFPredInterval}, but the intervals constructed under their approach are determined globally, e.g., by finding quantiles of the entire OOB error distribution. As a consequence, all prediction interval widths are uniform, lessening their utility in assessing the local uncertainty of individual predictions. However, unlike these intervals, the RF-FIRE intervals take advantage of known relationships between training and test data to estimate the local uncertainty of each out-of-sample prediction. The widths of the RF-FIRE intervals relate to the magnitude of their corresponding point prediction errors rather than remaining globally uniform for all predictions. 

To construct the intervals, we use the OOB residuals of the $k$ nearest training neighbors of each test instance. For a given test instance $x_i$, we denote the local distribution of OOB errors formed by the $k_i$-nearest training points as $\mathcal{D}^{k_i}$. The RF-GAP proximities constitute a similarity measure that accurately represents the random forest’s learned data geometry; consequently, these proximities are a feature-driven indication of which OOB residuals most closely match the distribution of possible errors for a new instance. The resulting residual distributions are functionally a Monte Carlo approximation derived from repeated bootstrap sampling of the random forest. This distribution can thus be used to perform inference and hypothesis tests, compare predicted errors, and construct prediction intervals.

The RF-FIRE prediction intervals follow intuitively by adding the empirical $(\alpha/2,1-\alpha/2)$ quantiles of the test instance's local residual distribution to the corresponding point prediction. With sufficiently large neighborhoods, these intervals match the desired $1-\alpha$ coverage level and often exhibit a narrower mean width than comparable methods~\cite{zhang2020RFPredInterval, meinshausen2006quantileforests} by exploiting known relationships between instances. 

\begin{figure*}
     \centering
     \includegraphics[width=0.55\linewidth]{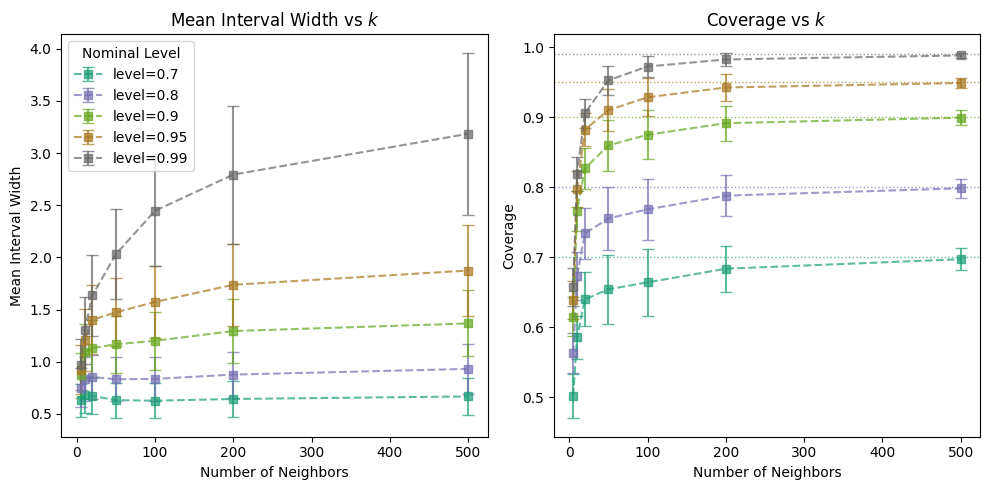}
     \caption{RF-FIRE interval widths and coverage across $k$ neighbors at different theoretical coverage levels: Average interval widths increase with $k$, outpacing convergence to the target coverage. This suggests the full distribution isn't needed for accurate coverage, and intervals remain narrower than those using the entire distribution. Changes in interval widths with increased $k$ are more pronounced at higher coverage levels.}
     \label{fig:coverage-width-by-k}
 \end{figure*}

For a given test data point $x_i$, the $(1 - \alpha) \times 100\%$ prediction interval for $y_i$ is given by  

\[
    \left[\hat{y}_i - \mathcal{D}^{(k_i)}_{\alpha/2}, \quad \hat{y}_i + \mathcal{D}^{(k_i)}_{1-\alpha/2} \right]
\]

where $\mathcal{D}^{(k_i)}_{\gamma}$ denotes the $\gamma$-th percentile of the distribution of OOB errors computed from the $k$ nearest neighbors of the test observation $x_i$. The number of neighbors $k_i$ can be determined dynamically for each test point by using all neighbors with a nonzero RF-GAP proximity to the test point. Alternatively, $k$ can be constant for all test instances and treated as a dataset-dependent hyperparameter chosen through cross-validation. Notably, the RF-FIRE prediction intervals are equivalent to the RF-Intervals in~\cite{zhang2020RFPredInterval} if $k = n$, the size of the training set. Thus, RF-FIRE can be viewed as a generalization of these RF-Intervals. Figure~\ref{fig:coverage-width-by-k} shows the tradeoff between coverage and mean interval width with an increase in $k$. Generally, the intervals are calibrated to the correct coverage levels using only about 200 neighbors (about 3\%, on average, of the OpenML-CTR23 dataset sizes)\footnote{The average number of observations for the OpenML-CTR23 is about 8200.} of the nearest neighbors, and the intervals are typically narrower than using the full OOB error distribution.

If $k$ is low relative to the number of training instances, the intervals depend only on local residuals and change in response to the magnitudes of their point predictions. Conversely, if $k$ equals the number of training observations, the intervals are a constant width for all observations and express only global uncertainty. Practitioners can choose the optimal balance between interval width and coverage, with lower $k$ values yielding narrower intervals on average. As the number of neighbors increases, the intervals grow wider and more globally uniform as they approach the guaranteed asymptotic coverage shown in~\cite{zhang2020RFPredInterval} using all training instances, although far fewer points are required.

\subsection{Classification - Trustworthiness Scores}\label{sec:classification}

Instance-based models, such as random forests or $k$-NN models, make predictions using a measure of proximity to other points. The certainty of the model prediction correlates with the homogeneity of the labels of nearby points. In regions of the decision space composed of a mixture of different classes, misclassification rates are higher, and even accurate model predictions can be viewed as less certain. 

One method to assess classification uncertainty is to take the difference in the top two predicted class probabilities for a given observation. Instead of computing a measure of heterogeneity among nearby points using class probabilities, we construct trust scores by determining the frequency of misclassification among nearby points using RF-GAP proximities to measure nearness. We call this method for classification uncertainty measurement Random Forest Instance Classification Evaluation trust scores, or RF-ICE, and describe two separate approaches to generate the scores.

The first approach combines information from OOB misclassifications and RF-GAP proximities. A vector of OOB misclassifications encodes the discrepancy between the true labels and the predictions made by the random forest model for each training point. By weighting these misclassifications via proximities, the RF generates trust scores ranging between 0 and 1 that represent the instance-level confidence.

We start by constructing an OOB misclassification vector and computing RF-GAP proximities. That is, given the vector

\[\mathbf{e} = [I(y_1 = \hat{y}^{\tilde{(B)}}_1), I(y_2 = \hat{y}^{\tilde{(B)}}_2), \cdots, I(y_n = \hat{y}^{\tilde{(B)}}_n)]^T \]
where $I$ is the $0-1$ indicator function, $\hat{y}_i^{\tilde{(B)}}$ is the OOB prediction for $x_i$, we define the RF-ICE trust scores as a proximity-weighted sum of OOB classifications, or 

\[
\mathcal{T}_i = \sum_{j = 1}^{n} w_{ij} I\left(y_j = \hat{y}^{\tilde{(B)}}_j\right) \rightarrow \mathbf{W} \cdot \textbf{e}
\]

We call this approach RF-ICE Expected Classification Rate (RF-ICE (ECR)). The ECR scores gauge the reliability of predictions based on the collective agreement of nearby training instances, as demonstrated by the accuracy of OOB predictions of points closely situated in the decision space. We note that a similar framework was established in~\cite{lu2021unified}, though these authors used a definition of OOB weight, rather than the RF-GAP proximities~\cite{rhodes2023rfgap}, which could induce bias in the results~\cite{tang2017rf-missdata}.

We define a second measure of trust based on the level of agreement between proximity values of points that belong to the same and opposing classes. We follow a similar approach to~\cite{devetyrov2010predswconfidence} in forming a measure of conformity as a ratio of proximity values. For a given point (test point or OOB), we calculate the ratio of $k$ highest proximities to points belonging to one class versus the $k$ highest proximities to points belonging to all other classes. For test points, we can calculate this ratio for each possible class and choose the class with the highest ratio as the predicted class. We call this method RF-ICE Conformity. We show that the RF-ICE Conformity measures better identify points that are likely misclassified. See results in Section~\ref{sec:results}. However, ECR presents more interpretable scores by forming an instance-level accuracy expectation.


\section{Experimental Setup and Results}\label{sec:results}

\begin{figure*}[!htb]
    \centering
    \includegraphics[width=.9\textwidth]{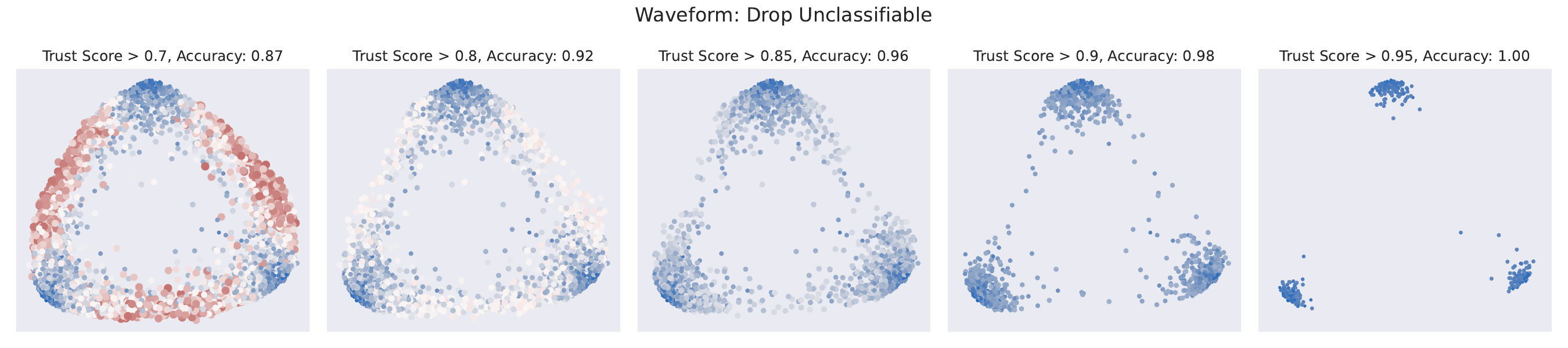}
    \caption{In each of the proximity-based embeddings~\cite{rhodes2021rfphate}, the points are sized and colored by their associated RF-ICE trust score. Red, large points indicate lower trust, while blue, smaller points have higher trust. We sequentially remove unclassifiable points according to the set thresholds of RF-ICE trust scores and the random forest model accuracy is recorded for each subset. The model accuracy improves as more unclassifiable points are removed.}
    \label{fig:wavewform-sequence}
\end{figure*}

For the regression task, we compare the interval coverage and widths of RF-FIRE with RF-Intervals~\cite{zhang2020RFPredInterval} and QRF~\cite{meinshausen2006quantileforests}. To assess overall performance, we devise a metric that balances the desired coverage levels and interval widths. Our evaluation uses all 35 datasets from the OpenML-CTR23 collection~\cite{fischer2023openmlctr}. All response variables were standardized to have zero mean and unit variance, and one-hot encoding was applied to datasets containing categorical variables.

For the classification task, we evaluate RF-ICE using the area under the accuracy-rejection curve~\cite{nadeem2009accuracy-rejection}. We compare against several nonconformity measures proposed in~\cite{devetyrov2010predswconfidence}, including prediction agreement across trees and differences in predicted class probabilities. Other random forest-based trust measures for classification were excluded due to a lack of available code. We conduct experiments on all 72 datasets from the OpenML-CC18 classification benchmark suite~\cite{oml-benchmarking-suites}. One-hot encoding was applied as needed, but no other preprocessing steps were performed.

Code to run all of the experimental results can be found at \href{https://github.com/JakeSRhodesLab/Localized-RF-Uncertainty}{https://github.com/JakeSRhodesLab/Localized-RF-Uncertainty}.

\subsection{Prediction Interval Results}

By default, we set the $k_i$ neighbors to be the number of points with non-zero proximity to $x_i$. If global coverage is more important for a particular use-case scenario, selecting $k \approx n$ will guarantee coverage, but the local influence will be diluted, and interval widths will become uniform. Empirically, we have seen that around 3-5\% of the dataset size is sufficient to attain the desired coverage.

We evaluated the trade-off between the two competing goals of producing narrow intervals and achieving accurate coverage. To assess the quality of prediction intervals, we use a score that balances both goals, which we call the Balanced Interval Score (BIS). The BIS penalizes both excessive width and deviations from a specified target coverage level. We define the BIS as

\[
\text{BIS} = \frac{1}{\text{width} \times \left(1 + \lambda \cdot \left| \text{coverage} - \text{target\_coverage} \right| \right)},
\]

where width is the length of the prediction interval, coverage is the empirical coverage rate, target\_coverage is the desired coverage level (e.g., 0.95), and $\lambda$ (represented as \texttt{penalty\_weight} in code) is a hyperparameter that controls how strongly deviations from the target coverage are penalized. This formulation ensures that BIS is high only when the interval is both narrow and well-calibrated, thereby aligning the scoring with practical desiderata for interval estimation.

To determine an appropriate value for $\lambda$ that equally balances the influence of interval width and coverage deviation in the BIS, we consider the scenario where a 10\% increase in interval width should have the same effect on the BIS as a 10 percentage point deviation from the target coverage.  Setting the BIS values equal in the two scenarios—one with a 10\% wider interval and perfect coverage, and the other with the original interval width but a 10\% deviation in coverage—we equate
\[
\frac{1}{1.1 \cdot w} = \frac{1}{w \cdot (1 + 0.10 \cdot \lambda)}.
\]
Solving this equation yields $\lambda = 1$, indicating that a value of 1 assigns equal importance to interval width and coverage deviation. Larger values of $\lambda$ place more emphasis on achieving accurate coverage, while smaller values prioritize narrower intervals. 

We calculated BIS scores across coverage levels of 70\%, 80\%, 90\%, 95\%, and 99\% and across all datasets. Aggregated results are found in Table~\ref{tab:BIS}. Here we see that RF-FIRE has the highest BIS scores on average, providing a better balance between width and coverage.

\begin{table}[!htb]
    \centering
    \caption{The BIS scores (higher is better) are averaged across all coverage levels and the 35 datasets of OpenML-CTR23. On average, RF-FIRE produced the best balance between coverage and interval width. $\lambda$ = 1 for these experiments.}
    \begin{tabular}{l c}
        \toprule
        Method & BIS ($\pm$ SEM) \\
        \midrule
        \textbf{RF-FIRE} & \textbf{1.971 $\pm$ 0.026} \\
        OOB Int. & 1.847 $\pm$ 0.015 \\
        QRF & 1.759 $\pm$ 0.012 \\
        \bottomrule
    \end{tabular}

    \label{tab:BIS}
\end{table}

\subsection{Classification Trust Results}

The classification trust scores indicate which points should be labeled as unclassifiable by the model, suggesting that these decisions may need reevaluation. To compare methods, we use accuracy-rejection curves, which assess classification methods as a function of their rejection rate~\cite{nadeem2009accuracy-rejection}. That is, as points are deemed unclassifiable according to a threshold value, they are no longer considered in the accuracy calculation. See Figure~\ref{fig:wavewform-sequence} for a visual reference denoting points that are difficult to classify. We use the area under the curves as a metric to assess the reliability of the classification trust assessment methods. In Table~\ref{tab:auc}, we compile results across all 72 OpenML-CC18 datasets and 10 random seeds. For each repeat, we calculate the accuracy-rejection AUC. Overall, RF-ICE Conformity provided the highest average AUC values. RF-ICE ECR outperformed differences in predicted probabilities and tree conformity. 

\begin{table}[!htb]
    \centering
    \footnotesize
    \caption{The accuracy-rejection AUC values were averaged across 10 random seeds and 72 datasets. Overall, RF-ICE Conformity had the highest scores.}
    \begin{tabular}{ll}
        \toprule
        Method & AUC \\
        \midrule
        \textbf{RF-ICE Conformity} & \textbf{0.965 ± 0.068} \\
        Conformity & 0.964 ± 0.069 \\
        RF-ICE (ECR) & 0.932 ± 0.099 \\
        Proba. Diff. & 0.929 ± 0.118 \\
        Tree Conformity & 0.730 ± 0.196 \\
        \bottomrule
    \end{tabular}
    \label{tab:auc}
\end{table}

\section{Conclusion}

In this paper, we introduced two novel random forest-based uncertainty measures for regression (RF-FIRE) and classification (RF-ICE) that use local similarity measures to estimate the expected errors of the test points by establishing a connection between them and the OOB errors of training points. We show that the RF-FIRE prediction intervals serve as a balance between guaranteed coverage with fixed interval widths, and acceptable coverage with more practical, locally-determined widths. We discussed a new method for generating classification prediction trust scores that employs RF-GAP proximities to associate each test prediction with the weighted accuracy of nearby training points via OOB prediction errors. Empirical results suggest that this approach is a valid means of assessing the uncertainty of a classification prediction. Using these methods, we can identify observations that are considered unclassifiable by the model and improve the overall accuracy by removing these points. RF-FIRE and RF-ICE empower users with practical tools to quantify the uncertainty of random forest predictions, improving the reliability and interpretability of these models in real-world applications. By integrating trust scores, users can make more informed decisions and enhance the transparency and interpretability of their decision-making process.

\bibliographystyle{IEEEtran}
\bibliography{main_abbrv}

\end{document}